\documentclass[letterpaper, 10 pt, conference]{ieeeconf}  

\IEEEoverridecommandlockouts                              

\makeatletter
\let\NAT@parse\undefined
\makeatother

\usepackage{subcaption}
\usepackage{graphics} 
\usepackage{graphicx}
\usepackage{grffile} 
\usepackage{amsmath} 
\usepackage{amssymb}  
\usepackage{color}
\usepackage{algorithm,algpseudocode}
\usepackage{mathtools}
\algnewcommand{\LeftComment}[1]{\Statex \(\triangleright\) #1}
\usepackage[T1]{fontenc}
\usepackage[utf8]{inputenc}

\usepackage{bm}
\usepackage{url}

\let\labelindent\relax
\usepackage{enumitem}
\usepackage[table,dvipsnames]{xcolor}
\usepackage[pagebackref=true,breaklinks=true,colorlinks,bookmarks=false]{hyperref}
\usepackage{graphicx,booktabs,array}
\usepackage[inkscapelatex=false]{svg}
\hypersetup{
linkcolor=BrickRed
,citecolor=Green
,filecolor=Mulberry
,urlcolor=NavyBlue
,menucolor=BrickRed
,runcolor=Mulberry
,linkbordercolor=BrickRed
,citebordercolor=Green
,filebordercolor=Mulberry
,urlbordercolor=NavyBlue
,menubordercolor=BrickRed
,runbordercolor=Mulberry
}

\title{\LARGE \bf
Towards Reinforcement Learning Controllers for Soft Robots using Learned Environments
}

\author{Uljad Berdica$^{1,2}$, Matthew Jackson$^{1,2}$, Niccolò Enrico Veronese$^{3}$, Jakob Foerster$^{1}$, Perla Maiolino$^{1,2}$
\thanks{$^{2}$ Supported by Autonomous Intelligent Machines and Systems (EPSRC Centre for Doctoral Training) EP/S024050/1 and EPSRC Programme Grant ‘From Sensing to Collaboration’ (EP/V000748/1)}
\thanks{$^{1}$University of Oxford
        {\tt\small \{uljad.berdica, matthew.jackson, perla.maiolino, jakob.foerster\}@eng.ox.ac.uk}}
\thanks{$^{3}$ Polytechnic University of Milan, Italy}
}

\begin{document}

\maketitle
\thispagestyle{empty}
\pagestyle{empty}

\begin{abstract}

Soft robotic manipulators offer operational advantage due to their compliant and deformable structures. However, their inherently nonlinear dynamics presents substantial challenges. Traditional analytical methods often depend on simplifying assumptions, while learning-based techniques can be computationally demanding and limit the control policies to existing data. This paper introduces a novel approach to soft robotic control, leveraging state-of-the-art policy gradient methods within parallelizable synthetic environments learned from data. We also propose a safety oriented actuation space exploration protocol via cascaded updates and weighted randomness. Specifically, our recurrent forward dynamics model is learned by generating a training dataset from a physically safe \textit{mean reverting} random walk in actuation space to explore the partially-observed state-space. We demonstrate a reinforcement learning approach towards closed-loop control through state-of-the-art actor-critic methods, which efficiently learn high-performance behaviour over long horizons. This approach removes the need for any knowledge regarding the robot's operation or capabilities and sets the stage for a comprehensive benchmarking tool in soft robotics control.\\

Code on \url{https://github.com/uljad/SoRoLEX}

\end{abstract}


\section{Introduction}\label{intro}

Soft robotic manipulators are made of compliant material and exhibit a low Young's modulus that enables them to be arranged in highly deformable geometries~\cite{laschi2016soft}. These designs, inspired by biological organisms, can undergo large elastic deformation throughout operations and facilitate safer interaction with the environments compared to their traditional rigid counterparts~\cite{thuruthuel2018control}. The morphological dexterity outsources parts of the solution computation to the compliant material~\cite{hauser2011towards}, but remains underactuated as the states of the physical body are governed by highly nonlinear continuum dynamics. Given the inherent challenges, the precise control of soft robots remains an open problem.

The existing analytical methods for accurate dynamic models in classical optimal control make reductive assumptions like constant-curvature and valve control heuristics for trajectory optimization~\cite{thuruthuel2018control} while relying on the material properties being unchanged or otherwise predictably modeled. These methods strive to reduce the computational complexity of the  dynamic model while not suppressing the modeling of the adaptive behavior that emerges through the soft robot's interaction with the environment. Moreover, parametric models fail to capture the computation embodied in the morphology of the robot, making data-driven models necessary for capturing the important insight from resulting deformations~\cite{laschi_learning-based_2023}.

Deep learning-based approaches utilize platforms with internal sensory data like pressure and Inertial Measurement Units (IMU)~\cite{gillespie2018learning} and external sensory data like visual trackers~\cite{thuruthel_model-based_2019} from the robot's interaction with the environment. This allows for the learned models to make use of the morphological changes that occur in operation time. However, the use of learned (black-box) mappings between actuation and task space comes at an increased computational cost both during training and at test time. The most commonly used architecture for such mapping is a non-linear autoregressive network with exogenous inputs (NARX) with one~\cite{thuruthel_model-based_2019} to four~\cite{thuruthel2017learning,alessi2023learning,pique2022controlling}~time delays. We also employ a recurrent architecture in the form of long short-term memory~(LSTM)~\cite{hochreiter1997long} as implemented in~\cite{flax2020github} for faster training and inference time. Both NARX and LSTM architectures are designed to learn from distant interactions by overcoming the gradient vanishing problem. While NARX networks tackle this problem through delayed connection from distant past, the contribution is small and scales the computation by a factor equal to that of time-delayed connections~\cite{dipietro2017analyzing}. Considering entire sequences of actuations and observations is essential for learning closed loop control of soft robots that is not privy to and limited by prior knowledge of their dynamics.

Reinforcement Learning (RL) algorithms have been successful in solving sequential decision making problems under these limitations by learning through repeated interactions with the environment which in this work is represented as a recurrent model trained from collected data on the robot. Popular success stories include Proximal Policy Optimization (PPO)~\cite{schulman2017proximal}, which has been particularly successful in continuous control~\cite{lillicrap2015continuous,van2012reinforcement}. The adoption of deep learning based methods in RL, their large computational requirements, and algorithmic complexity has caused an explosion of different frameworks. These frameworks aim to balance high performance hardware utilization and ease-of-use for rapid prototyping~\cite{hessel2021podracer}. A recent breakthrough is PureJaxRL~\cite{lu2022discovered}, which uses JAX~\cite{bradbury2018jax} to run agents and environments jointly on the GPU, resulting in order of magnitude speedup compared to prior approaches. 

\begin{figure*}[ht]
    \centering
     \includegraphics[width=0.9\textwidth]{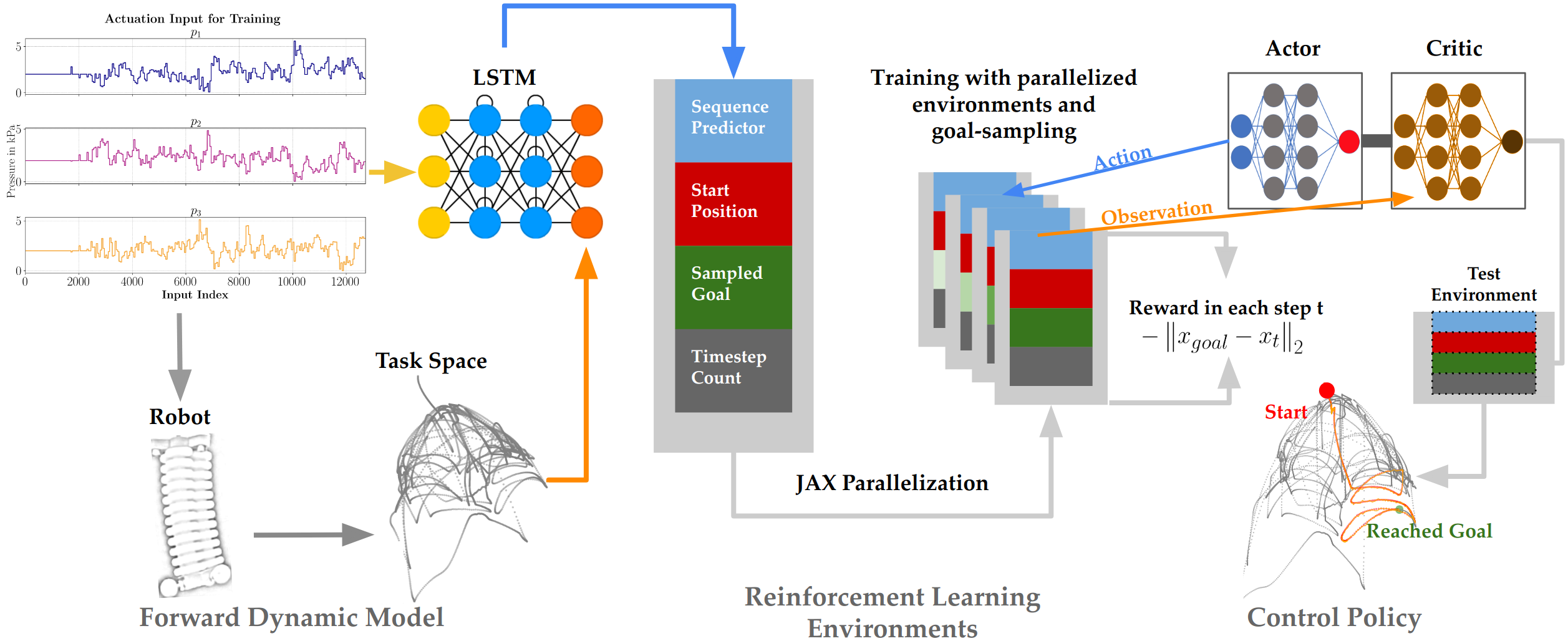}
    \caption{The pipeline of the learned environment-based solution proposed in this work. The recurrent network to the left represents the LSTM at the core of the synthetic environments. }
    \label{pipeline}
    \vspace{-5mm}
\end{figure*}

Previous work on the use of RL for the control of soft includes value-based methods in~\cite{satheeshbabu2019open,you2017model} and with early attempts of actor-critic methods in~\cite{ansari2017towards}. The most recent work to date using closed loop policy gradients leverages Cosserat Rod Models simulations to learn the forward dynamics~\cite{alessi2023learning}. These methods hinge on the limited number of recorded interactions in the dataset,  prior knowledge of material heuristics~\cite{satheeshbabu2019open} or of guiding trajectories for the policy search~\cite{rolf2010goal,thuruthel_model-based_2019}. We propose a methodology based on forward dynamic models learned in absence of simulations or guiding trajectories.

This work seeks to bring together the advantages of learning-based solutions to soft robotics control by utilizing SOTA PPO implementations~\cite{lu2022discovered} to learn closed loop controllers inside environment models implemented via high-performance computing libraries~\cite{lange2022gymnax}. Our method bypasses the need for any analytical models or prior information regarding the robot's operation. The forward dynamics model is learned by training with the data collected on the robot through a mean reverting random walk in actuation space. Feedforward and recurrent control policies are learned by interacting with the parameterized forward dynamics model wrapped in a JAX-based environment. 

In this paper, we first introduce the methods used to collect that data from the robot with examples of the generated input sequences. We then describe the architecture and training process for the supervised learning of the forward dynamics model through a sequence-to-sequence (seq2seq) prediction model as well as the policy optimization procedure to train the the actor-critic network. Finally, we present the training results and inference examples of the closed-loop policies conditioned on observations and tested in the forward dynamic model.


\section{Methods}\label{sec:methods}

Figure~\ref{pipeline} shows the full pipeline implemented in this work starting from the state space exploration to the left~(~\ref{subsec:babbling}), the creation of the dataset~(\ref{data_prep}) and the training of the forward dynamic model~(\ref{subsec:forward}) using a LSTM network. We leverage JAX to generate parallel environments~\ref{subsec:RLEnv} from the trained LSTM model and actor-critic network to simultaneously learn from multiple training trajectories sampled directly from a task space distribution to update policies without depending on previous example trajectories(~\ref{subsec:RLexplain}). This method enables batching multiple goals for robust policy learning. Finally consecutive goal learning is shown by selecting a desired target position for which an action sequence is generated.

\subsection{State Space Exploration}\label{subsec:babbling}
The sequence-to-sequence (seq2seq) mapping from actuation space to task space requires sufficient exploration to reproduce a representative environment interaction for the online actor-critic training. The exploration in this work does not require any static workspace assumptions and is exclusively in the actuation space.

The single constraint to this exploration policy is the maximum pressure $P_{max}$ allowed across the valves of the robot to ensure the full functionality and structural integrity of the soft robot. The protocol to cover a wide range of robot configurations by applying $p_j$ actuation pressures within the safety value uses a mean reverting random walk~\cite{WolframMeanReverting2011} with tunable parameters, specifically a sigmoid scaling ${1}/({1+e^{-x}})$ with two parameters $\alpha$ and $\beta$ to account for the randomness and the position of the sigmoid inflection point respectively. $\alpha$ weights the previous input's effect on the next one whereas $\beta$ modulates the average value of the total pressure. Each generated term $p_{i+1}^*$ of each valve is incrementally added to the previous pressure $p_{i}^*$ with the generated terms being normally distributed around the preloaded pressure value $p_b$ of the robot in the resting state. Eq.~\ref{exploration_equation} describes the cascading update procedure to keep the sum of $p_j$ below $P_{max}$ in every iteration i.

 \begin{equation}
\begin{aligned}
\forall i \in \{0, 1, \ldots, N\}, \  \forall j \in \{0, 1, \ldots, N_{valves}\}   \\
p_{i+1,j}^{*} = \alpha \ p_{i,j} + (1-\alpha) \ \mathcal{N}({p_b,1})\\
p_{i+1} = P_{max} \ \sigma \left(\beta \ \sum_{j}^{}p_{i,j}^{*}\right) \ \frac{p_{i+1,j}^{*}}{\sum_{j}^{}p_{i,j}^{*}}
\end{aligned}
\label{exploration_equation}
\end{equation}
\begin{itemize}
    \item $i$ indexes the iteration,
    \item $j$ indexes the valves,
    \item $N$ is the total number of iterations,
    \item $N_{valves}$ is the total number of valves.
\end{itemize}

\subsection{Forward Dynamic Model Learning}\label{subsec:forward}
We use an LSTM to learn the dynamic mapping between the actuation and task space. This has been chosen to be able to train for significantly longer sequences. To increase the predictive versatility of the learned environment around the LSTM latent states, the testing pairs are generated by using a sliding window approach with a step of one in permuted order. Each training pair in the dataset contains the three actuation pressures and the corresponding robot reference point in Cartesian coordinates. Each sequence consists of 512 steps. This number of steps was chosen as it is on average 100 steps higher than the mean reverting random walk procedure which allows to record the initial preload pressure state and the return to that initial state upon the end of the exploration. The length of these exploratory runs is dependent on the limitations of physical platforms used in section~\ref{subsec:robot}.

\begin{figure}[h]
    \centering
    \includegraphics[width=0.9\columnwidth]{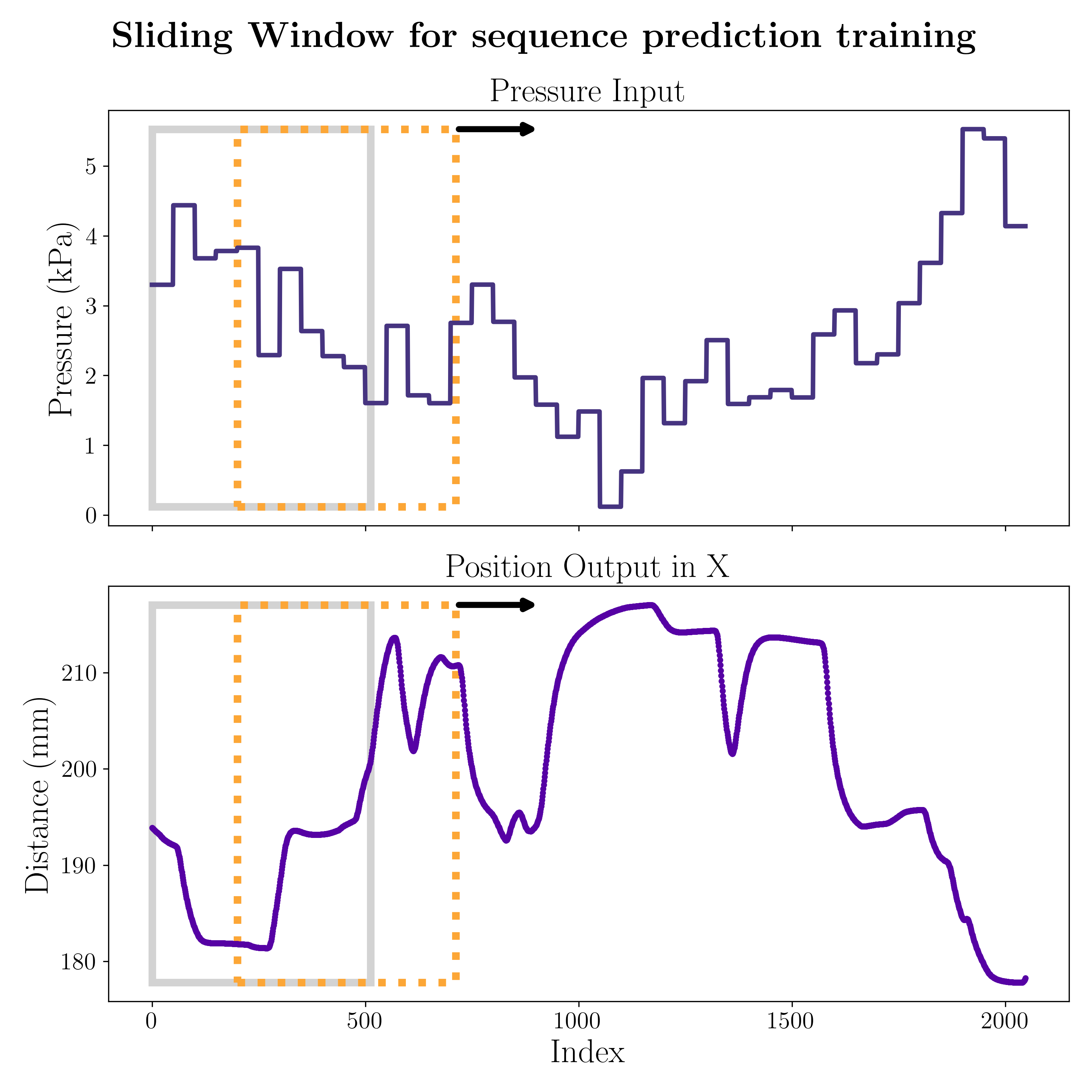}
    \caption{Training Pair Generation is shown through matched colors. 
    For illustration purposes we use a sequence length of 512 with \textit{step size} of 200 on a subset of the data. In practice, we use a \textit{step size} of 1 and slide through the entire runs.}
    \label{sliding_window}
    \vspace{-2mm}
\end{figure}

This is illustrated in Fig.~\ref{sliding_window} for the x-direction and the actuation pressure $p_1$ where the training pairs are shown with matched colors. The sliding sequence method effectively places every data points in every possible context of the sequence during training time which allows for context-independent prediction of outputs and the learning of behaviors that were not seen in random exploration like reaching unseen targets as the results in sec~\ref{sec:results} show. 

The model was trained by dividing the dataset in training and testing with a 75-25\% split. To improve the generalisation capability of the model, we consider two approaches for the order of passing the data to the model in training. One is to feed the generated sequences as they appear in the dataset to maintain as much of the history of the system as possible. The other approach is to randomly permute the order in which the sequences are passed through the network meaning that temporally close sequences can be seen by the training network at time steps that are further apart than the time difference of their occurrence in the data. The empirical observations in section~\ref{sec:training_order_results} show that the random permutation of the sequence pairs performs better with test sequences that were not seen during training. 

\subsection{Generation of Reinforcement Learning Environment}\label{subsec:RLEnv}
The RL agent only interacts with multiple instances of an environment learned from an offline dataset.
The environments in this paper are chosen as platforms to interface with the learned dynamic model in a more structured way that allows the policy network to be trained for different goals in each episode that are sampled from the task space i.e. Cartesian coordinates range of the collected data.

As mentioned in sec~\ref{intro}, we leverage JAX~\cite{bradbury2018jax} to achieve high-performance training and inference from our learned environment. JAX objects are compiled with XLA and executed in parallel on GPU. This enables both our model and policy to be trained entirely on GPU, leading to huge speedups against CPU-based methods. We implement our model in the Gymnax environment framework \cite{lange2022gymnax}, allowing existing agent implementations for online RL to be used seamlessly with our model. The Gymnax frameworks implements environments as classes the instances of which can be run in parallel and have different states allowing for different training conditions and rewards.


\subsection{Policy Optimization}\label{subsec:RLexplain}

We used a policy-gradient methods to train an actor-critic network~\cite{hessel2021podracer,schulman2017proximal}. 
Specifically, we use proximal policy optimization (PPO) \cite{schulman2017proximal} for policy optimization. PPO uses a trust region approach to stabilise the policy gradient update, employing a clipped surrogate objective which prevents the parameterized policy diverging too far from its original value after it is updated. Policy methods are preferred over value-based method for this robotics application due to the constrained policy update when optimizing for an objective function. 

The policy is trained by concatenating the final target to the observations from the environment. If the target destination in task space has been reached within the measurement error distance of 1 mm or if the episode terminates after a predetermined number of steps, the environment resets to the initial state and samples from a range within the reachable dynamic space of the robot before starting the new training episode steps. Note that this is not necessarily a target included on the dataset.

The implementation of the goal perturbation is done via the Algorithm~\ref{goal_perturbation_algo}. The algorithm takes in the Forward Dynamic Model, the parameters of the environment like starting position, total time steps per episode as well as the training configuration with the values of the relevant hyperparameters like number of updates, batch size and learning rate. A new goal is set every time the environment state is reset. The output is the parameters of the conditional probability function of an action for a given observation, also referred to as a policy $\pi( \cdot | observation)$ in reinforcement learning terms. Policy $\pi$ is obtained by training the actor-critic network that learns to predict the actions and values of observations or embedding of observations depending  on whether the network is feed-forward or recurrent respectively. Both these networks are trained using PPO~\cite{schulman2017proximal} implementations based on PureJaxRL~\cite{lu2022discovered}.

\begin{algorithm}
\caption{Policy Training with Goal Perturbation}
\begin{algorithmic}[1]
\renewcommand{\algorithmicrequire}{\textbf{Input:}}
\renewcommand{\algorithmicensure}{\textbf{Output:}}
\Require \\Forward Dynamic Model,\\ Environment Parameters,\\ Training Configuration
\Ensure Parameters of control policy $\pi(\cdot | observation)$
 \LeftComment{Initialisation of network and environment parameters}
\State $env\_params \leftarrow$ Environment Parameters
\State $\pi \leftarrow$ initialize actor-critic network parameters
\State $train\_config \leftarrow $ Training Configuration
\LeftComment create parallel $environment$ with Gymnax~\cite{lange2022gymnax}
\State $environment \leftarrow$ Gymnax($env\_params$)
\LeftComment{Loop condition values from Training Configuration}
\For{$t < \texttt{TOTAL\_UPDATES}$ \textbf{or} not converged} 
    \If {$t = 0$}
        \State $observation \leftarrow$ initial observation
        \State $perturbation \leftarrow env\_params$
        \State $initial\_pose \leftarrow env\_params$
        \State $goal \leftarrow \mathcal{N}(0,1) \cdot perturbation + initial\_pose$
    \EndIf
    \State $actions \leftarrow \pi(observation)$
    \State $observation \leftarrow$ Forward Dynamic Model$(actions)$
    \State $reward \leftarrow -\left \lVert goal - observation \right \rVert_2$
    \State $\pi \leftarrow$ PPO$(reward, goal, train\_config)$~\cite{lu2022discovered}
    \State $t \leftarrow t + 1$
\EndFor  
\Ensure $\pi(\cdot | observation)$
\end{algorithmic}
\label{goal_perturbation_algo}
\end{algorithm}



\section{Experimental Setup}\label{section:physical_setup}

\subsection{Robot}\label{subsec:robot}
\begin{figure}[ht]
    \centering
    \includegraphics[width=0.9\columnwidth]{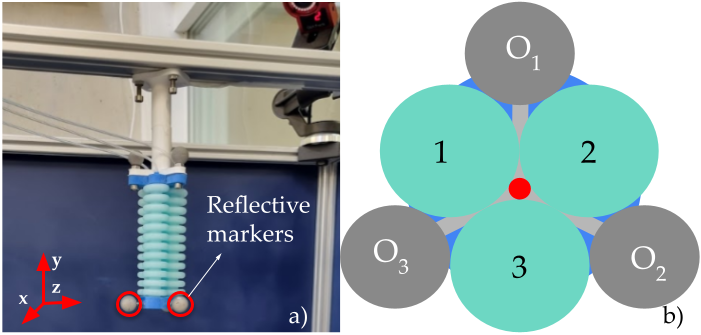}
    \caption{Robot at initial positions (a) no deformation at home position with initial baseline pressure 2kPa, (b) Transverse cross-section view of the root, pressure chambers A to C and reflective markers $O_1$ to $O_3$}
    \label{robot}
    \vspace{-2mm}
\end{figure}

To validate our approach we use a soft a three chambers bellow-shaped actuator connected to a rigid frame in Fig.~\ref{robot}. The actuation is achieved with compressed air controlled by three separate proportional  Festo valves (VEAA-L-3-D2-Q4-V1-1R1). The number of controllable inputs corresponds to the number of chambers, therefore, $N_{valves}$ is 3~(see eq.~\ref{exploration_equation} in section~\ref{subsec:babbling}). Reflective markers are integrated at the top and bottom of the actuator to track its movements. The position of the robot reference point in Cartesian coordinates is estimated as the centroid of the triangle marked by the bottom grey reflective markers in~Fig.~\ref{robot} using an Optitrack Motive system equipped with four Flex 3 Cameras. The valves are controlled within a ROS2~\cite{macenski2022robot} environment.

\subsection{Dataset Preparation}\label{data_prep}

\begin{figure}[h!]
    \centering
    \includegraphics[width=0.85\columnwidth]{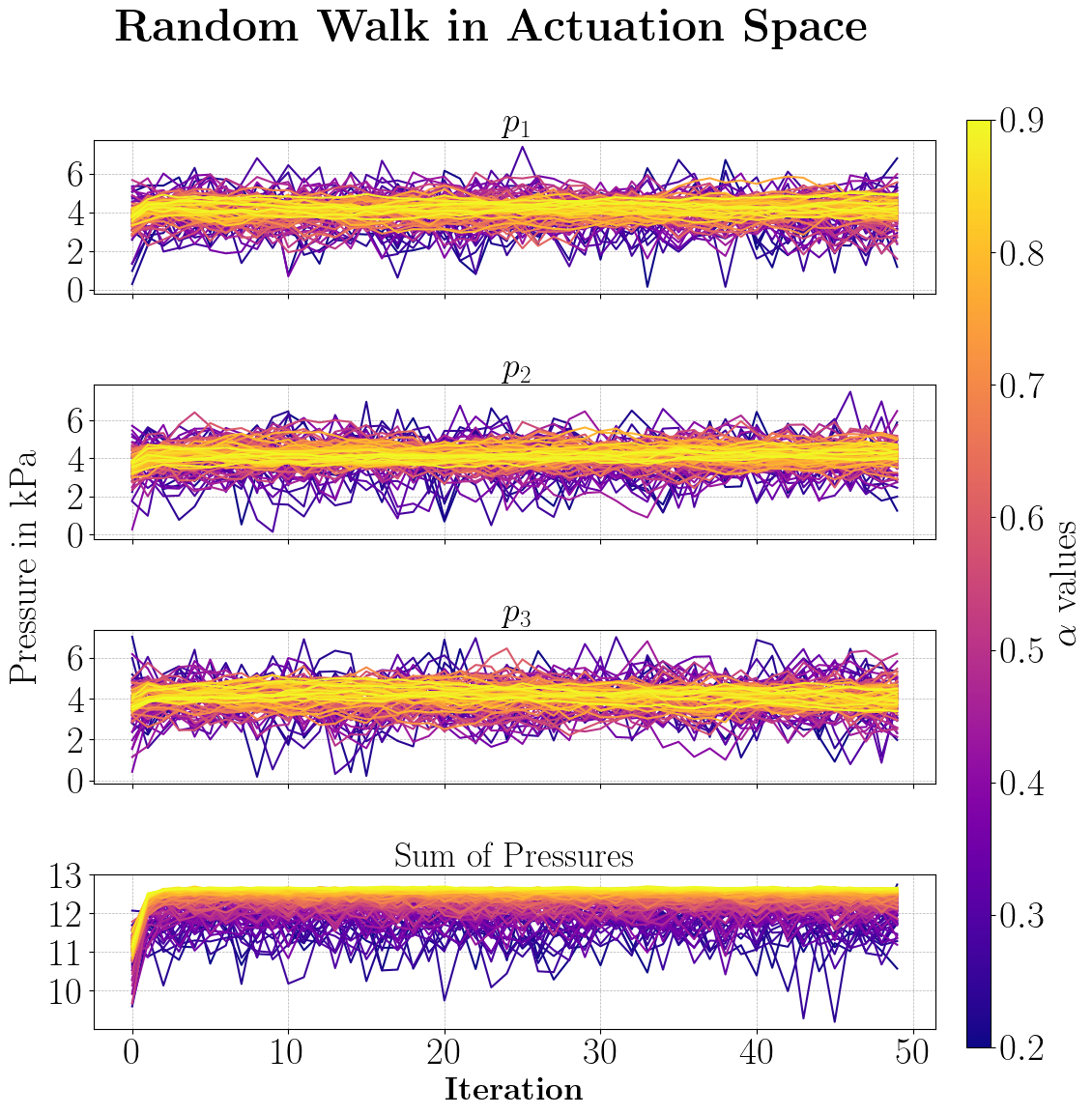}
    \caption{Random Walk in actuation space for different exploration hyperparameter $\alpha$ }
    \label{input_explore}
    \vspace{-2mm}
\end{figure}

The mean reverting random walk under the pressure constraints from section~\ref{subsec:babbling} is implemented with a  $P_{max}$ of 13 kPa, a $p_b$ of 2kPa, $N_{valves}$ of 3 and $N$ of 50. Fig.~\ref{input_explore} shows the generated input pressure action sequences for different values of $\alpha$. An $N$ of 50 iterations was chosen for clarity of visualization to demonstrate that the inputs are independent and diverse while not exceeding the safety limit of $P_{max}$.

\begin{figure}[h]
    \centering
    \includegraphics[width=0.85\columnwidth]{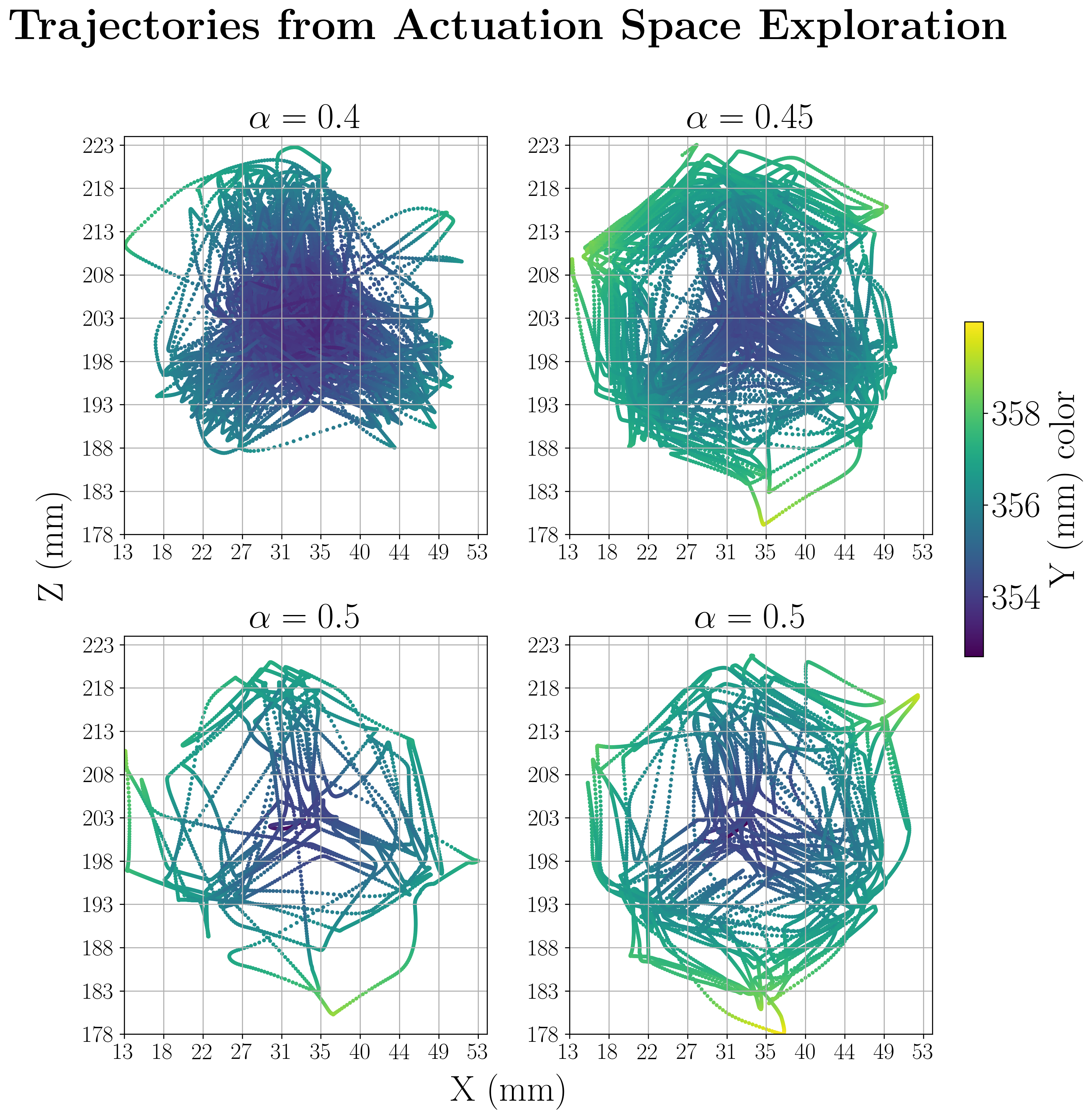}
    \caption{Resulting trajectories from a random walk in actuation space various levels of randomness}
    \label{output_explore}
    \vspace{-2mm}
\end{figure}

Fig.~\ref{output_explore} shows the resulting task space trajectories as a result of the exploration method implemented for different $\alpha$ values of randomness. 

The control pressure data is collected through the ROS2 interface and the consecutive positions of the markers are collected with the Optitrack. The Optitrack measurements are acquired at a frequency higher than that of the pressure measurements. To make sure every measured Cartesian coordinate corresponds to a pressure measurement for the training with the sequences of pairs described in~\ref{subsec:forward}, we match every pressure value to the nearest position in time through a nearest neighbors search through the timestamps of each Optitrack measurement. The rows in the Optitrack measurements that do not get matched to a pressure value are filled with the earliest possible pressure value as the pressure in the robot remains the same before it is changed in a step-like manner. The control frequency is kept at a constant of 2Hz to accommodate to the physical limitations of the setup and enable the covering of the whole dynamic motion range of the robot.


\section{Results}\label{sec:results}

\subsection{Forward Dynamic Model}\label{sec:training_order_results}
\begin{figure}[h]
    \centering
    \includegraphics[width=0.85\columnwidth]{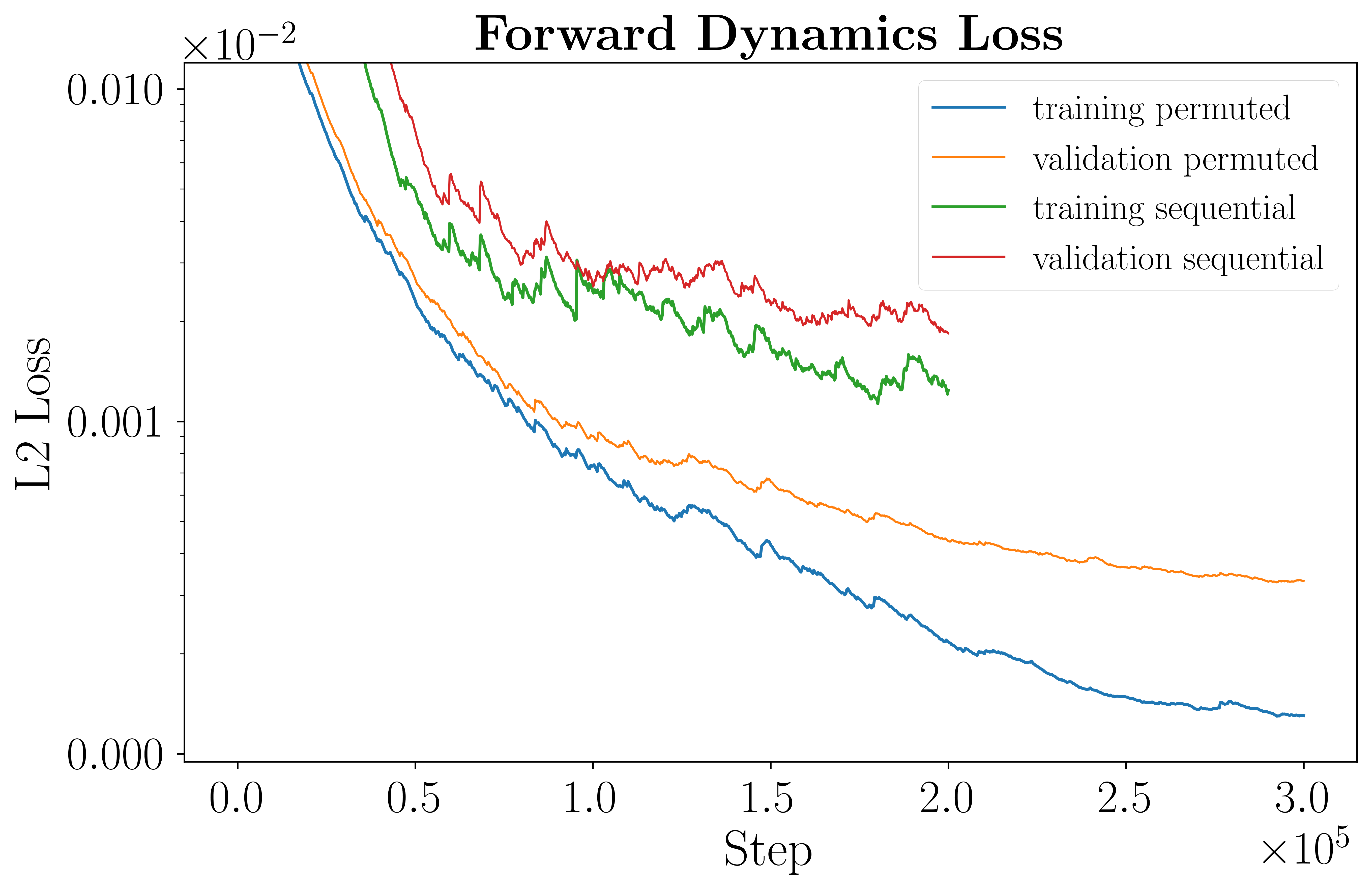}
    \caption{Training and Validation Losses for the Forward Model. The plots are smoothed using an exponential moving average with a factor of 0.025 }
    \label{loss_curve} 
    \vspace{-2mm}
\end{figure}

Fig.~\ref{loss_curve} shows the training and test results for the mapping from actuation to cartesian positioning via the collected data. The two approaches to passing the training dataset through the model described in section~\ref{subsec:forward} were implemented and tested on the same set of sequences held out during training. In $2 \cdot 10^5$ training steps, it becomes clear that the general approach of permuting the order of temporally close sequences achieves higher train and test results than the sequential training with the same sequences.

\subsection{Task Space Reconstruction}

From Fig.~\ref{loss_curve} we can see that both training and validation losses for the randomly permuted case are low showing that the model accurately predict the task space. However, it is necessary to evaluate the ability of the forward dynamic model to reconstruct a correct task space from exploratory input sequences in test time as it is a core requirement for the data efficient development of closed loop control policies via learned environments. 

\begin{figure}[h]
    \centering
    \begin{subfigure}[b]{1\columnwidth}
        \includegraphics[width=0.8\linewidth]{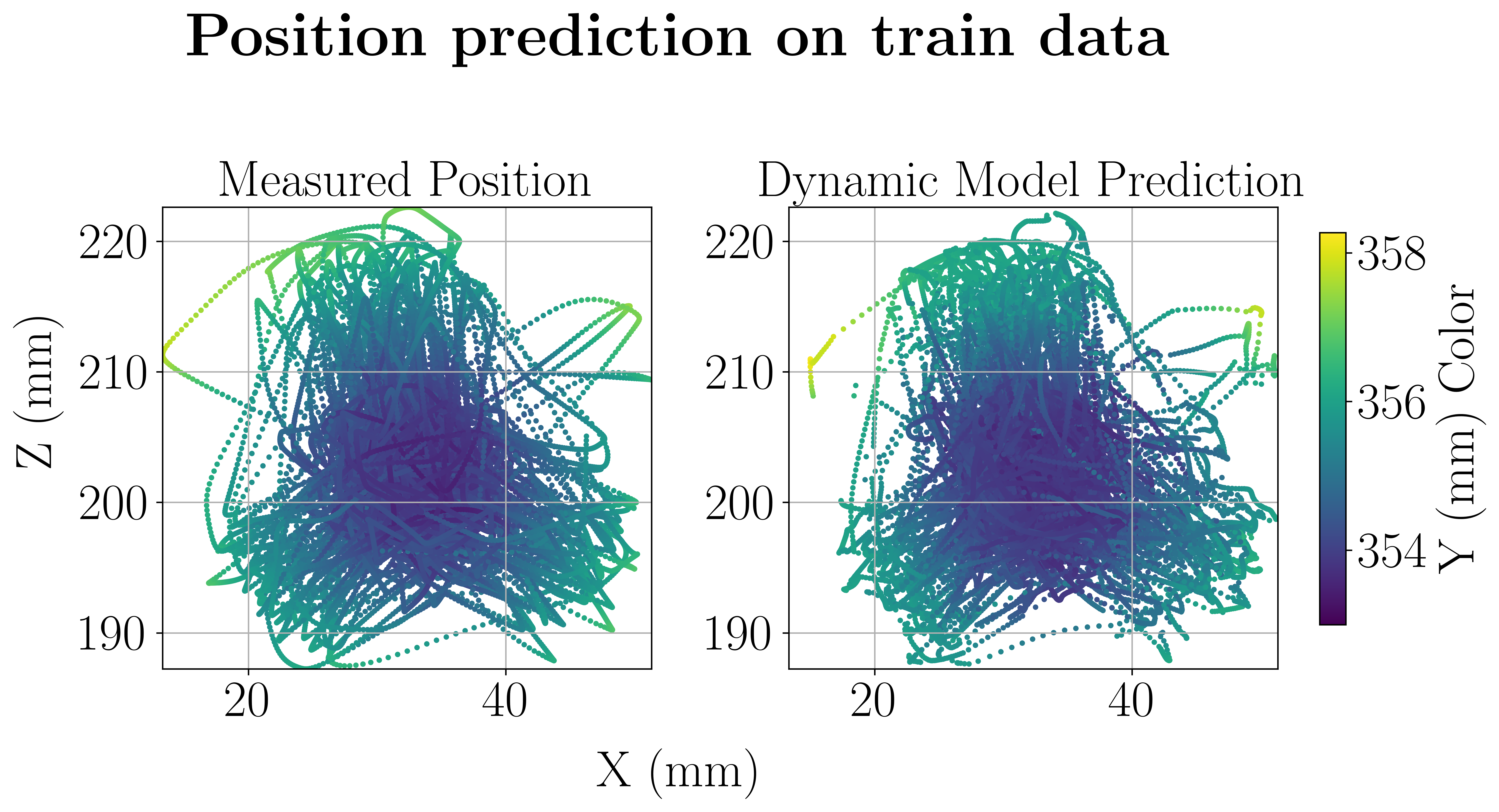}
        \caption{Reproduction of training data as seen on top left of Fig.~\ref{input_explore}}
        \label{fig:dynamic_train}
    \end{subfigure}
    \vspace{0.1em} 
    
    \begin{subfigure}[b]{1\columnwidth}
        \includegraphics[width=0.8\linewidth]{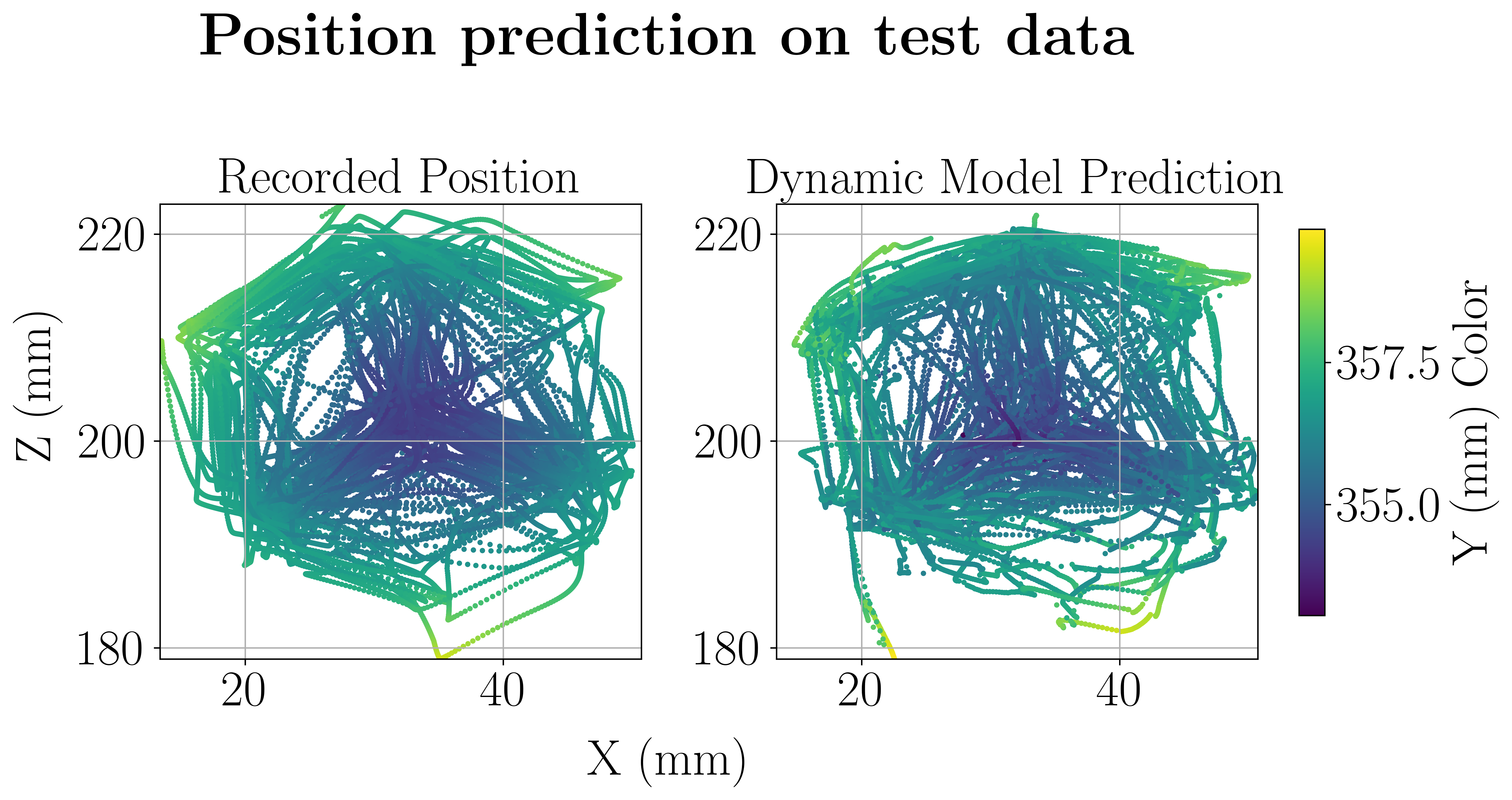}
        \caption{Reproduction of unseen action space exploration sequence}
        \label{fig:dynamics_test}
    \end{subfigure}
    \caption{Predictive performance of the model on training and testing datasets.}
    \label{fig:combined_dynamics}
    \vspace{-2mm}
\end{figure}

We evaluate this aspect by qualitatively comparing the task space reconstruction from data seen during the training and data from previously unseen sequences. Fig.~\ref{fig:dynamic_train} shows a close reconstruction of training data and Fig.~\ref{fig:dynamics_test} shows a close reconstruction of an entire run that has not been used in training. These results set the ground for the use of learned models in RL environments.

\subsection{PPO on the Learned Environment}

To evaluate the impact of the robots movements history in the episode we implemented two different policies. Specifically one conditioned to the latest observation and one conditioned to the entire history of observations (recurrent). 

\begin{figure}[h]
    \centering
    \includegraphics[width=0.8\columnwidth]{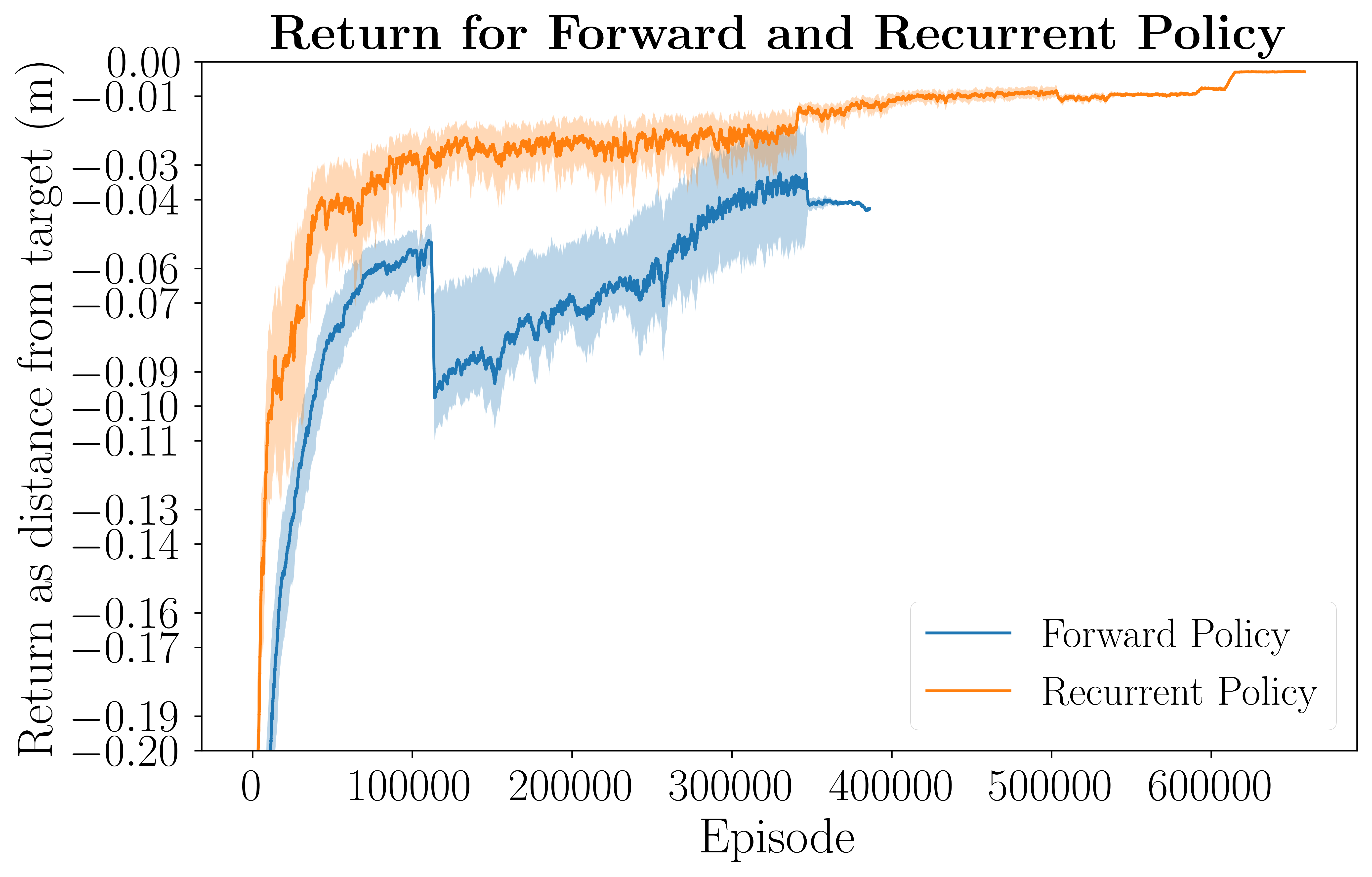}
    \caption{Episodic return for Forward and Recurrent Policies with one standard deviation shaded region for 20 different random seeds}
    \label{random_seed_both}
    \vspace{-2mm}
\end{figure}

Fig.~\ref{random_seed_both} shows the rewards monotonically increase with the number of episodes. This also further validates this result as the rewards is monotonically increasing to convergence within 3mm of the target. The mean with one standard deviation shaded region obtained by running with 20 different random seeds is plotted for each policy type. 

\section{Conclusion}\label{sec:conclusion}

In this paper, we use a model-free approach to learn the forward dynamics of a soft robotic arm. We develop a protocol for state space exploration using random walks and use the generated data to train our model. We demonstrate the effectiveness of our approach in recreating tasks from test sequences and show its potential for developing closed-loop control policies in soft robotics. This general methodology for developing a closed loop control policies shows great promise towards establishing new soft robotics control benchmarks and bridging the physical advantages of soft robotics with the most recent work in Machine Learning. The demonstrated ability of synthetic environments to facilitate planning on real-world data can provide a path towards future work in data-driven emergence of complex behavior, learned sim2real adaptation strategies and further testing of such policies on physical robots with generated actuation regimes beyond general exploration.

\bibliographystyle{IEEEtran}

\bibliography{IEEEabrv,BibFiles/RL,BibFiles/Robotics}

\end{document}